\documentclass[letterpaper]{article} % DO NOT CHANGE THIS
\usepackage{booktabs}
\usepackage{aaai25}  % DO NOT CHANGE THIS
\usepackage{times}  % DO NOT CHANGE THIS
\usepackage{helvet}  % DO NOT CHANGE THIS
\usepackage{courier}  % DO NOT CHANGE THIS
\usepackage[hyphens]{url}  % DO NOT CHANGE THIS
\usepackage{graphicx} % DO NOT CHANGE THIS
\usepackage{natbib}  % DO NOT CHANGE THIS AND DO NOT ADD ANY OPTIONS TO IT
\usepackage{amsmath}
\usepackage{caption} % DO NOT CHANGE THIS AND DO NOT ADD ANY OPTIONS TO IT
\usepackage{algorithm}
\usepackage{algorithmic}
\usepackage{amssymb}

\usepackage{newfloat}
\usepackage{listings}
\DeclareCaptionStyle{ruled}{labelfont=normalfont,labelsep=colon,strut=off} % DO NOT CHANGE THIS
\floatstyle{ruled}
\newfloat{listing}{tb}{lst}{}
\floatname{listing}{Listing}
\title{State Estimation and Control of Dynamic Systems from High-Dimensional Image Data}
\author {
    Ashik E Rasul\textsuperscript{},
    Hyung-Jin Yoon\textsuperscript{}
}
\affiliations {
    \textsuperscript{}Department of Mechanical Engineering,\\Tennessee Technological University, Cookeville, TN, USA\\
    arasul42@tntech.edu, hyoon@tntech.edu
}

\usepackage{bibentry}
\begin{document}

\maketitle

\begin{abstract}
Accurate state estimation is critical for optimal policy design in dynamic systems. However, obtaining true system states is often impractical or infeasible, complicating the policy learning process. This paper introduces a novel neural architecture that integrates spatial feature extraction using convolutional neural networks (CNNs) and temporal modeling through gated recurrent units (GRUs), enabling effective state representation from sequences of images and corresponding actions. These learned state representations are used to train a reinforcement learning agent with a Deep Q-Network (DQN). Experimental results demonstrate that our proposed approach enables real-time, accurate estimation and control without direct access to ground-truth states. Additionally, we provide a quantitative evaluation methodology for assessing the accuracy of the learned states, highlighting their impact on policy performance and control stability.
\end{abstract}

% Uncomment the following to link to your code, datasets, an extended version or similar.
%
% \begin{links}
%     \link{Code}{https://aaai.org/example/code}
%     \link{Datasets}{https://aaai.org/example/datasets}
%     \link{Extended version}{https://aaai.org/example/extended-version}
% \end{links}

\begin{figure*}[t]
\centering
\includegraphics[width=0.9\textwidth]{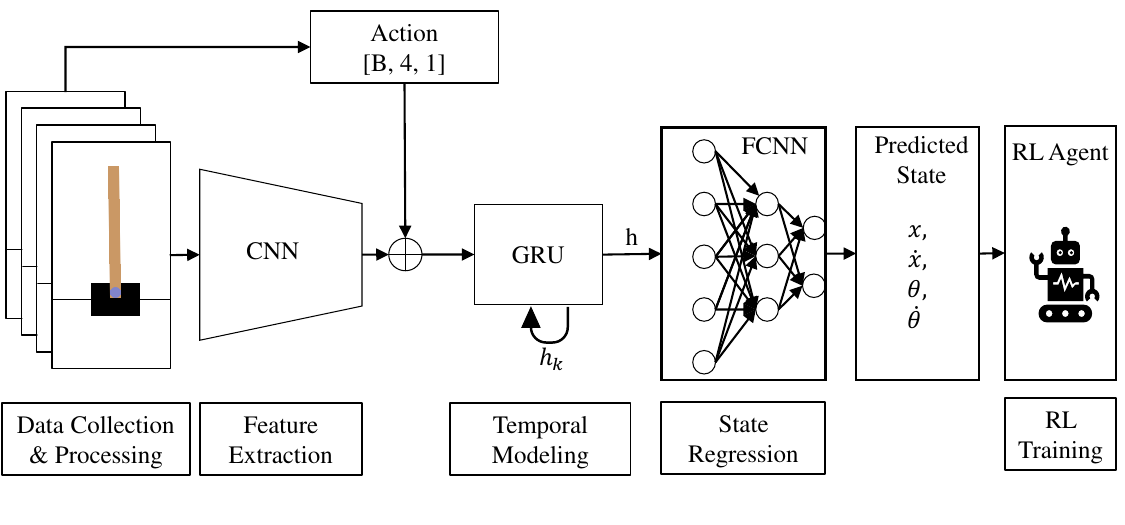} % Reduce the figure size so that it is slightly narrower than the column.
\caption{Proposed framework for dynamic system state prediction and RL training}
\label{framework}
\end{figure*}

\section{Introduction}

Autonomous systems operating in real-world environments, such as unmanned aerial vehicles (UAVs) and autonomous vehicles, frequently rely on actionable state information extracted from high-dimensional sensory data like images or videos. Humans possess a natural perceptual capability to interpret sequential visual cues and infer temporal properties such as velocity. However, replicating this ability in artificial systems involves two key challenges: extracting important spatiotemporal features from high-dimensional data and estimating the true state of the system from noisy observations.

Deep Neural Networks (DNNs) have consistently demonstrated superior performance over traditional computer vision methods~\citep{grigorescu2020survey} in perception-based tasks for autonomous systems. Convolutional neural networks (CNNs) are widely used for extracting hierarchical spatial features from images~\citep{lecun2015deep}. To capture sequential dependencies in visual data, recurrent neural networks (RNNs), especially gated recurrent units (GRUs), are widely adopted due to their effectiveness in modeling temporal relationships with relatively lightweight architectures~\citep{cho2014learning}. Leveraging the universal function approximation capability of fully connected neural networks (FCNNs)~\citep{hornik1991approximation}, the latent representations produced by GRUs can be mapped to interpretable, low-dimensional state predictions of dynamic systems, bridging the gap between high-level perception and actionable insights required for optimal control.

Conventional deep reinforcement learning (RL) algorithms, such as Deep Q-Networks (DQN), often assume full access to the environment’s true state vector. While this assumption simplifies policy learning, it does not reflect real-world conditions, where direct measurement of all states is typically expensive or challenging. In such systems, the true state values must instead be estimated through processing one or multiple sensor inputs.

This research addresses these challenges by proposing a perception-based controller design framework (as shown in Figure~\ref{framework}) that integrates spatiotemporal feature extraction with learned state estimation. Our framework employs a CNN to extract spatial features from individual image frames. These features, combined with corresponding actions, are processed through a GRU to capture temporal dependencies. An FCNN subsequently maps these learned representations to physically meaningful, low-dimensional state estimates. By processing sequences of frames and incorporating action history, the model implicitly captures latent dynamics like velocity, filtering out irrelevant information and emphasizing task-relevant features.

The estimated states serve as inputs to a reinforcement learning (RL) agent to train an optimal policy. To the best of our knowledge, this framework represents the first approach explicitly designed to extract interpretable state information for optimal controller design from raw visual inputs. Our key contributions are summarized as follows:

\begin{itemize}
\item We develop a perception-based reinforcement learning framework\footnote{\url{https://github.com/arasul42/cartpoleStatePred}} that extracts interpretable state predictions from high-dimensional image data.
\item We demonstrate that an RL agent trained on these predicted states achieves comparable performance to an RL agent trained using full state information.
\item We present a quantitative evaluation methodology to independently assess state prediction accuracy and controller performance.
\end{itemize}

\section{Background}

Perception-based control relies heavily on compressing high-dimensional observations into low-dimensional representations that preserve task-critical temporal features. To process grid-structured data such as images, Convolutional Neural Networks (CNNs) are most widely used. They utilize convolutional layers with learnable filters to extract local and hierarchical features from raw input data, followed by non-linear activation functions like ReLU to introduce complexity into the model. CNNs with backpropagation were successfully applied to identify handwritten ZIP codes~\citep{lecun1998gradient}. However, when the number of layers in the neural network is increased, traditional backpropagation becomes challenging. They face problems such as local optima, gradient vanishing, gradient exploding or overfitting. The introduction of ImageNet and multi-hidden-layer pretraining ~\citep{krizhevsky2012imagenet} shows the capability of CNNs in large scale image classification tasks. However, despite their effectiveness in capturing spatial patterns, standard CNNs are limited in modeling long-range temporal dependencies in sequential data, as they lack an explicit mechanism to capture the order or temporal dynamics inherent in time-series. This shortcoming motivates the use of recurrent architectures for tasks where temporal context plays a critical role.

Traditional Fully Connected Neural Networks (FCNNs) demonstrated high accuracy in regression tasks~\citep{chemali2018state} but suffer from an explosion in the number of parameters when processing long sequences. Recurrent Neural Networks (RNNs) inherently handle sequential data better and have demonstrated superior performance compared to FCNNs. However, they face gradient vanishing and exploding issues for long input sequences~\citep{bengio1994learning}. To overcome these limitations, Gated RNNs, such as Long Short-Term Memory (LSTM) and Gated Recurrent Units (GRU), introduce gating mechanisms to regulate information flow and capture long-term dependencies. LSTMs have shown exceptional performance in various domains, including time series forecasting~\citep{siami2019performance}, machine translation~\citep{sutskever2014sequence}, and healthcare diagnostics~\citep{balaji2021automatic}. However, their complex architecture and high computational cost hinder real-time deployment. To balance computational efficiency and accuracy, GRUs were developed, offering comparable performance with a simpler structure~\cite{yang2019state}, making them more suitable for real-time applications.

Reinforcement learning (RL) is a branch of machine learning in which agents learn to make decisions by interacting with an environment to maximize cumulative rewards~\citep{sutton1998reinforcement}. Q-learning is a widely used RL algorithm that enables agents to learn a Q-function, which estimates the expected return of taking a particular action in a given state. The introduction of the Deep Q-Network (DQN)~\citep{mnih2015human} combined Q-learning with deep neural networks, allowing agents to learn control policies directly from high-dimensional sensory inputs such as raw pixel images. DQN incorporates key innovations such as experience replay—where the agent’s experiences are stored and randomly sampled during training—and a target network, which is updated periodically to stabilize learning and prevent divergence.
Conventional implementations of Deep Q-Networks (DQN) typically assume access to full state information~\citep{stable-baselines3}. However, in real-world scenarios, such complete observability is often unattainable due to sensor limitations or occlusions. To address this, reinforcement learning based on low-dimensional latent representations has gained significant traction~\citep{lee2020stochastic}. Despite their success, the lack of interpretability in these latent-based approaches poses a challenge for deployment in safety-critical or high-stakes applications.

\section{Proposed Framework}
Our proposed framework as illustrated in Figure~\ref{framework} has following components:

\subsubsection{Data Acquisition \& Preprocessing \\ }
\begin{figure}
    \centering
    \includegraphics[width=0.95\linewidth]{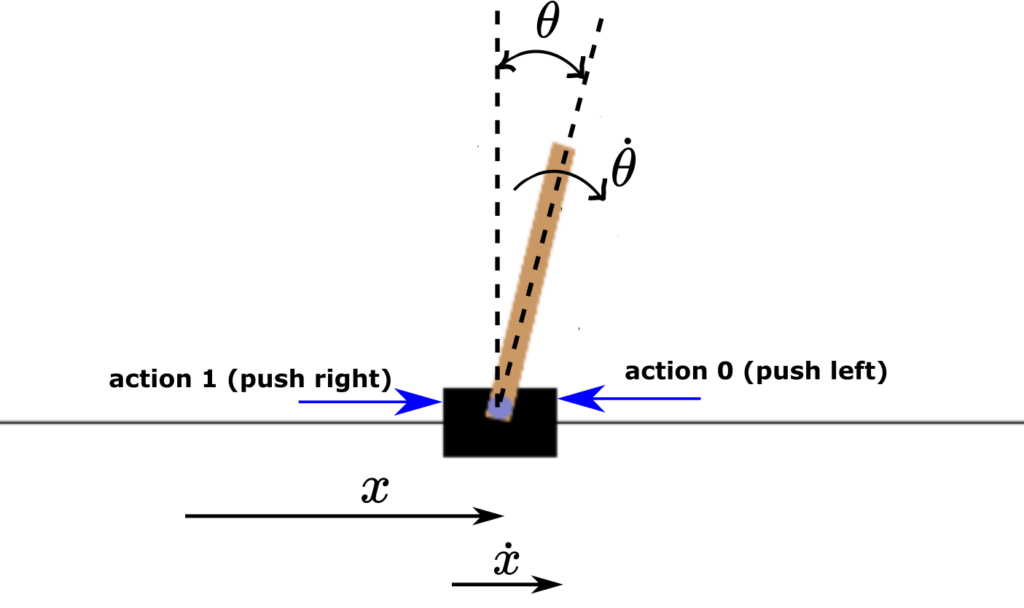}
    \caption{CartPole Environment in OpenAI Gym~\cite{Haber2023}}
    \label{fig:cartpole}
\end{figure}

We use the OpenAI Gym simulation environment~\cite{brockman2016openai} as the data generation source for our framework. The system consists of four state variables ($s_k$) at timestep $k$: the cart's position $(x)$, the cart's linear velocity $(\dot{x})$, the pole's angular position $(\theta)$, and the pole's angular velocity $(\dot{\theta})$ as shown in Figure~\ref{fig:cartpole}. The action space comprises two discrete actions: action 0 moves the cart to the left, while action 1 moves it to the right. The agent's objective is to take actions that keep the pole within a bounded angular range$([-0.21,0.21])$ and the cart within a bounded displacement$([-2.4,2.4] )$ as illustrated in Figure~\ref{fig:cartpole}. The agent collects reward $(r=1)$ for each timestep. Each episode is truncated at $500$ timesteps if not terminated earlier by going out of bounds defined for $x$ and $\theta$.

\begin{figure}
    \centering
    \includegraphics[width=0.85\linewidth]{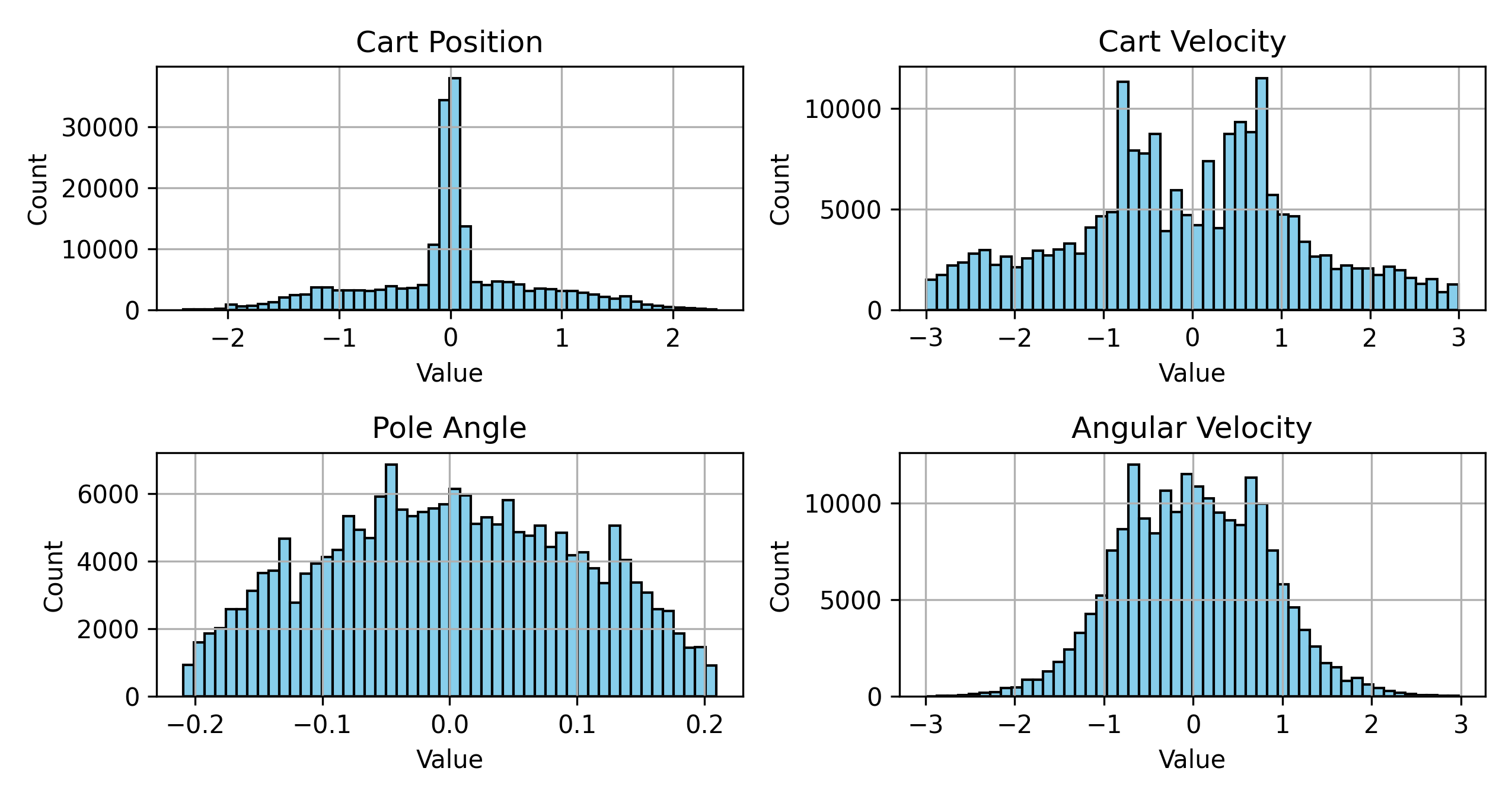}
    \caption{Dataset distribution across the state space}
    \label{fig:dataset}
\end{figure}

To generate the training samples, the agent applies random actions on the environment. To ensure balanced representation across the state space, we discretize the continuous CartPole state dimensions into bins and perform environment resets to generate samples from each bin as shown in Figure~\ref{fig:dataset}. The dataset is collected at a frame rate of 30 FPS with an image resolution of \(128 \times 128\) pixels using the RGB rendering mode of the simulator. Each rendered frame is resized and normalized by scaling pixel values to the range \([0, 1]\) through division by $255$. These preprocessed RGB image sequences, along with the corresponding action labels, are used as input for training our framework.

\subsection{Convolutional Feature Extraction with CNN}
We feed a sequence of 4 RGB image frames \(\{x_k\}_{k=1}^4\) through the CNN encoder to extract spatial features. Each image frame \(x_k \in \mathbb{R}^{3 \times 64 \times 64}\) is passed through two convolutional layers with ReLU activations and downsampling by a factor of 2 at each layer. This produces a compact feature map, which is then flattened and passed through a fully connected layer to produce a 128-dimensional feature vector:
\begin{equation}
    f_k = \text{CNN}_\phi(x_k), \quad f_k \in \mathbb{R}^{128},
\end{equation}
where \(\phi\) denotes the parameters of the frame encoder. 

\subsection{Temporal Modeling with GRU}
We use a Gated Recurrent Unit (GRU) to capture temporal dependencies across the sequence of image frames and corresponding control actions. The GRU receives a sequence of 4 inputs \( \{p_k\}_{k=1}^{4} \), where each \( p_k = [f_k; a_k] \) is a concatenation of the convolutionally encoded image feature \( f_k \in \mathbb{R}^{128} \) and the scalar action \( a_k \in \mathbb{R} \) at timestep \( k \).

Following the formulation of \cite{chung2014empirical}, the hidden state \( h_k \in \mathbb{R}^{d_h} \) at each timestep is computed as a convex combination of the previous hidden state \( h_{k-1} \) and the candidate hidden state \( \tilde{h}_k \), modulated by the update gate \( z_k \):

\begin{equation}
h_k = (1 - z_k) \odot h_{k-1} + z_k \odot \tilde{h}_k
\end{equation}
The update and reset gates are computed as:
\begin{equation}
z_k = \sigma(W_z [h_{k-1}, p_k])
\end{equation}
\begin{equation}
r_k = \sigma(W_r [h_{k-1}, p_k])
\end{equation}
The candidate hidden state \( \tilde{h}_k \) is defined by:
\begin{equation}
\tilde{h}_k = \tanh(W_h [r_k \odot h_{k-1}, p_k])
\end{equation}

\begin{figure}
    \centering
    \includegraphics[width=.95\linewidth]{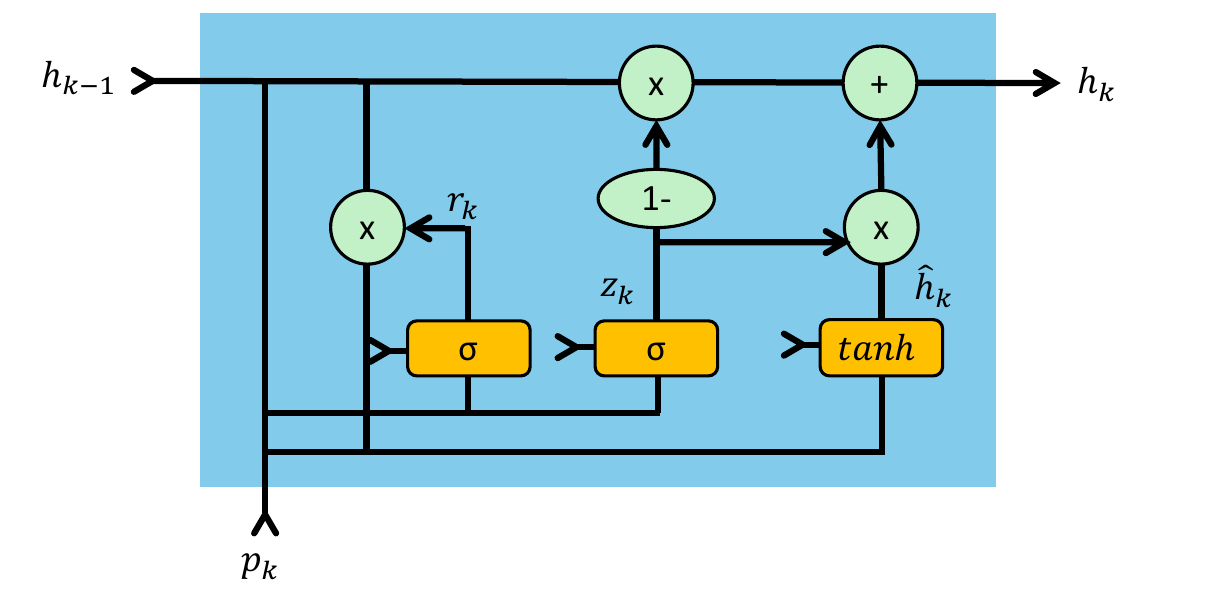}
    \caption{Gating mechanism of the GRU}
    \label{fig:gru}
\end{figure}

Here, \( \sigma(\cdot) \) denotes the sigmoid activation function, \( \tanh(\cdot) \) is the hyperbolic tangent function, and \( \odot \) represents element-wise multiplication. The weight matrices \( W_z, W_r, W_h \) are learnable parameters of the GRU. The gating mechanism of GRU is illustrated in Figure~\ref{fig:gru}. After processing the 4-timestep sequence \( \{p_1, p_2, p_3, p_4\} \), the final hidden state \( h_4 \) serves as a compact representation of the spatiotemporal history of observations and actions.

\subsection{State Prediction with FCNN}
The final hidden state \( h_4 \) is passed through a fully connected neural network (FCNN) to predict the system's physical state \( \hat{s}_{k+4} = [x, \dot{x}, \theta, \dot{\theta}] \):
\begin{equation}
    \hat{s}_{k+4} = \phi_{\text{fcnn}}(h_4),
\end{equation}
where \( \phi_{\text{fcnn}} \) denotes the parameters of the regression head. The model is trained end-to-end to minimize the mean squared error (MSE) between the predicted and ground truth states:
\begin{equation}
    \mathcal{L} = \left\| \hat{s}_{k+4} - s_{k+4} \right\|^2
\end{equation}

\subsection{Reinforcement Learning with DQN}
If the estimated state is close to the true state, then the next state depends only on the current estimated state and action. Given data of \((\hat{s}_k, r_k, \hat{s}_{k+1}, a_k)\), we can estimate the action-value function. The Bellman optimality equation for the action-value function is:
\begin{equation}
    Q^*(\hat{s}_k,a_k) = r(\hat{s}_k,a_k) + \gamma \mathbb{E} \left[ \max_{a'} Q^*(\hat{s}_{k+1}, a') \right]
    \label{dqn_eqn}
\end{equation}
where \( Q^*(\hat{s}_k,a_k) \) is the optimal action-value function, \( r \) denotes the reward, \( \gamma \) is the discount factor, and \( \hat{s}_{k+1} \) is the next estimated state.

\subsection{Evaluation Metrics}
Evaluation focuses on both state estimation accuracy and RL state tracking performance with respect to a reference trajectory.

\subsubsection{State Estimation Accuracy:}  
Given the true state \( s_k = [x_k, \dot{x}_k, \theta_k, \dot{\theta}_k] \) and the estimated state \( \hat{s}_k \), prediction accuracy is evaluated using the Root Mean Squared Error (RMSE) over \( T \) timesteps:
\begin{equation}
    \text{RMSE} = \sqrt{\frac{1}{T} \sum_{k=1}^T \left\| s_k - \hat{s}_k \right\|^2}
\end{equation}
To report relative accuracy, the RMSE values are normalized by the respective state variable bounds and expressed as percentages.

\subsubsection{RL Policy Tracking Error:}  
Let the desired reference state be \( s_{\text{ref}} = [0, 0, 0, 0] \). The Mean Absolute Error (MAE) between the actual controlled state \( s_k \) and the reference trajectory is computed as:
\begin{equation}
    \text{MAE} = \frac{1}{T} \sum_{k=1}^{T} \left| s_k - s_{\text{ref}} \right|
\end{equation}
As with RMSE, the MAE values are also normalized by the respective state variable bounds and reported as percentages.

\section{Experiments and Results}
\subsection{Training of State Prediction Model}
\begin{figure}
    \centering
    \includegraphics[width=0.85\linewidth]{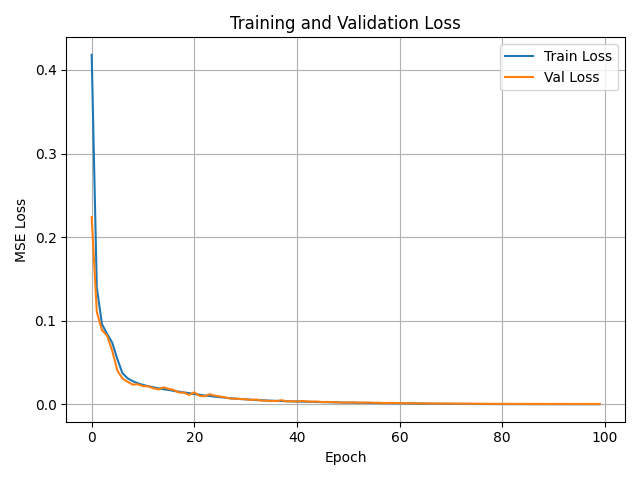}
    \caption{Training and validation loss of the prediction model}
    \label{fig:pred_loss}
\end{figure}

We collect and preprocess a dataset of size \( 200{,}000 \). The dataset is randomly split into training and validation sets with an 80:20 ratio. We use a batch size of 32 and train the model for 100 epochs with a learning rate of \( \text{lr} = 1 \times 10^{-3} \). We observe that both training and validation losses converge after 60 epochs, as shown in Figure~\ref{fig:pred_loss}.

\begin{figure}
    \centering
    \includegraphics[width=0.85\linewidth]{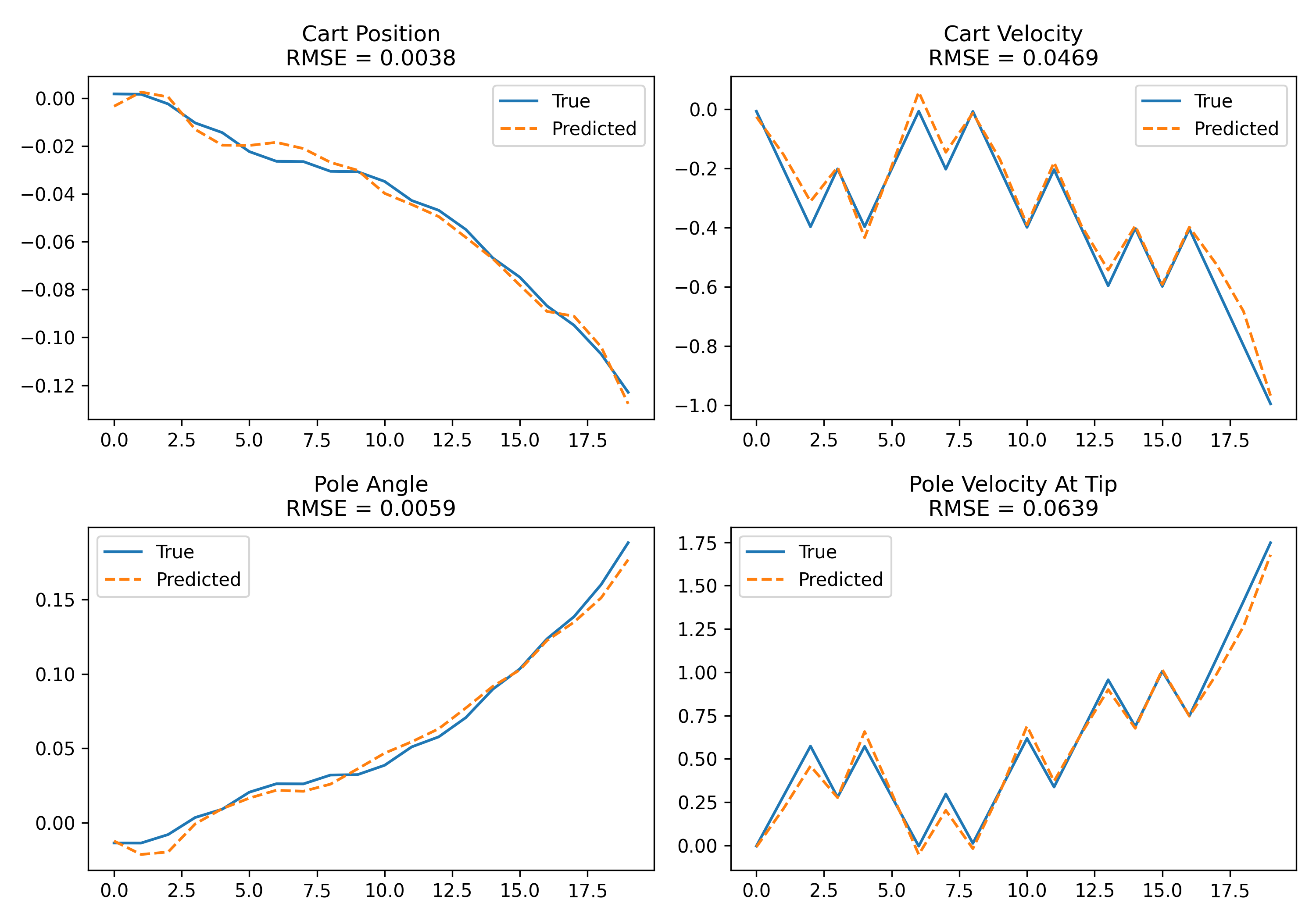}
    \caption{Evaluation of the state prediction model on a single episode}
    \label{fig:one_step_evaluation}
\end{figure}

To test the learned model, we generate test image frames by applying random actions in the CartPole environment. The image frames are recorded until the episode terminates due to a high pole angle (\( \theta = 0.2 \)). Although initial prediction errors are low as shown in Figure~\ref{fig:one_step_evaluation}, the simulation timespan is too short to draw definitive conclusions. To evaluate the prediction model over a longer time horizon and measure the tracking error, we train two separate reinforcement learning (RL) models: one using full state access, and the other using the predicted state from our model.

\subsection{Reinforcement Learning with Full State Observation}

We first establish a reinforcement learning (RL) baseline assuming full observability of the system. Under this assumption, the agent has access to the true state vector \( s_k = [x_k, \dot{x}_k, \theta_k, \dot{\theta}_k] \) at each timestep. We use the Deep Q-Network (DQN) algorithm, where the agent learns an approximation \( Q(s, a; \theta) \) as defined in Equation~\ref{dqn_eqn}. The temporal-difference (TD) error is given by:
\begin{equation}
    \delta = r_k + \gamma \max_{a'} Q(s_{k+1}, a'; \theta') - Q(s_k, a_k; \theta)
\end{equation}

We minimize the Huber loss over a batch of experiences:
\begin{equation}
    \mathcal{L} = \frac{1}{B} \sum_{i=1}^{B} \mathcal{L}(\delta_i), \quad \text{where } \mathcal{L}(\delta) =
    \begin{cases}
        \frac{1}{2} \delta^2 & \text{if } |\delta| \leq 1 \\
        |\delta| - \frac{1}{2} & \text{otherwise}
    \end{cases}
\end{equation}

We adopt an \( \epsilon = 2\% \)-greedy exploration strategy, where the action is selected randomly with probability \( \epsilon \) and greedily with probability \( 1 - \epsilon \). The exploration rate decays by 10\% over time to encourage early exploration and later exploitation.

\begin{figure}
    \centering
    \includegraphics[width=0.85\linewidth]{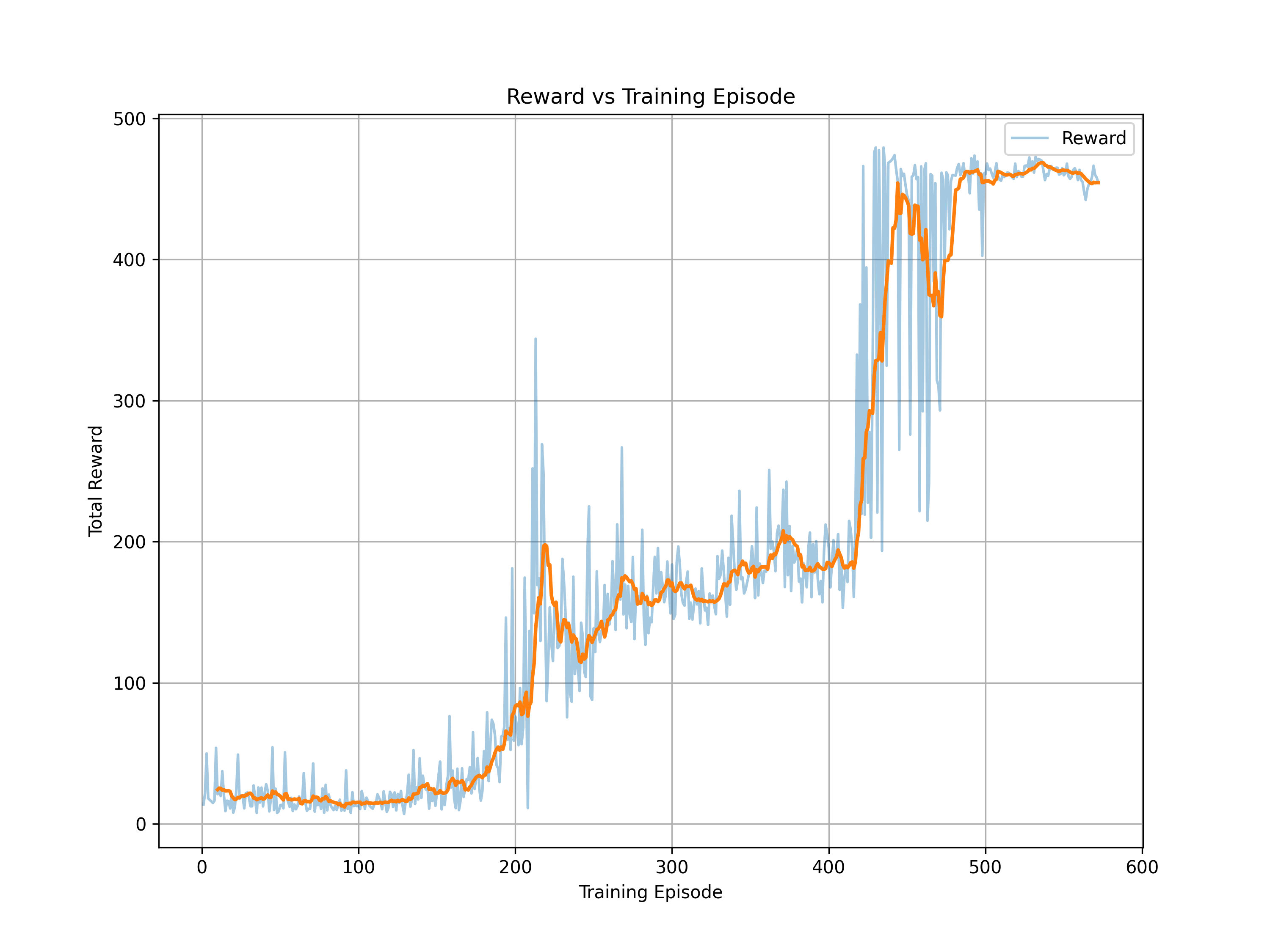}
    \caption{DQN training with full state observation}
    \label{fig:reward_plot}
\end{figure}

To enhance data efficiency, we use an experience replay buffer of size 100{,}000, which stores past transitions and samples minibatches uniformly during training.

The default reward function does not encourage the cart to be at the center or pole angle to be at the minimum. To encourage smoother and centered control, we augment the default reward with penalties on cart displacement, pole angle deviation, and action switching (jerk), using weights \( \lambda_1 = 0.1 \), \( \lambda_2 = 1 \), and \( \lambda_3 = 0.35 \), respectively. The updated reward function becomes:
\begin{equation}
    r_k = 1 - \lambda_1 x_k - \lambda_2 \theta_k - \lambda_3 \Delta u_k
\end{equation}

We train the RL model with a learning rate \( \text{lr} = 1 \times 10^{-4} \), discount factor \( \gamma = 0.99 \), batch size of 64, and for 100{,}000 timesteps. Each episode is truncated at 500 timesteps if not terminated earlier. We observe steady reward convergence after 450 episodes, as illustrated in Figure~\ref{fig:reward_plot}.

\begin{figure}[t]
    \centering
    \includegraphics[width=0.85\linewidth]{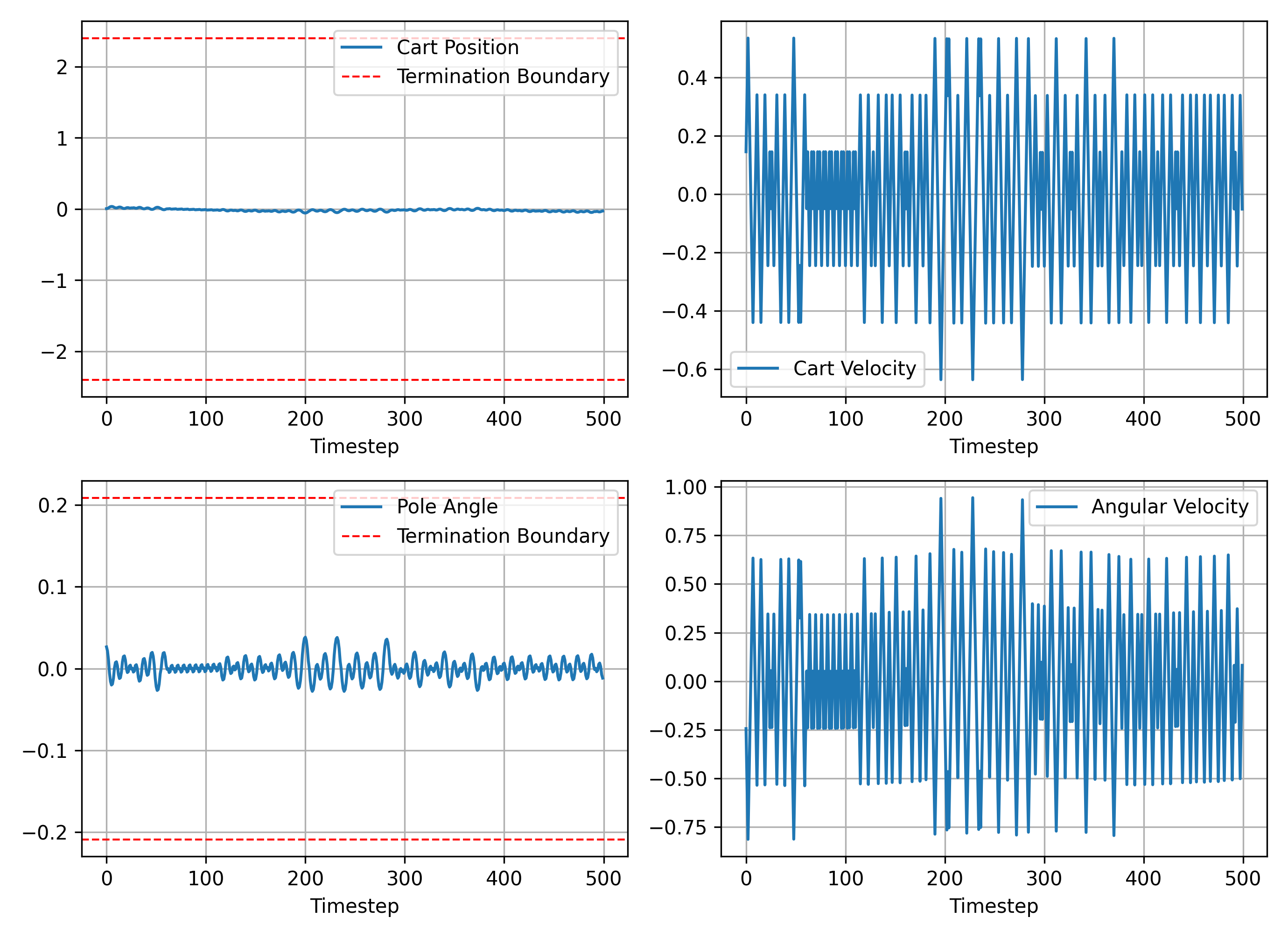}
    \caption{DQN performance with true state observation}
    \label{fig:rel_full}
\end{figure}

We evaluate the trained DQN agent in the same environment with access to the full state. The policy achieves stable control with low tracking errors. This evaluation serves as a performance benchmark for the dynamic system. The trajectory tracking is shown in Figure~\ref{fig:rel_full}, an illustrative video is available here\footnote{\url{https://youtu.be/7gsOLwh1Ei0}}.

\subsection{Reinforcement Learning with State Prediction}

\begin{figure}[t]
    \centering
    \includegraphics[width=0.85\linewidth]{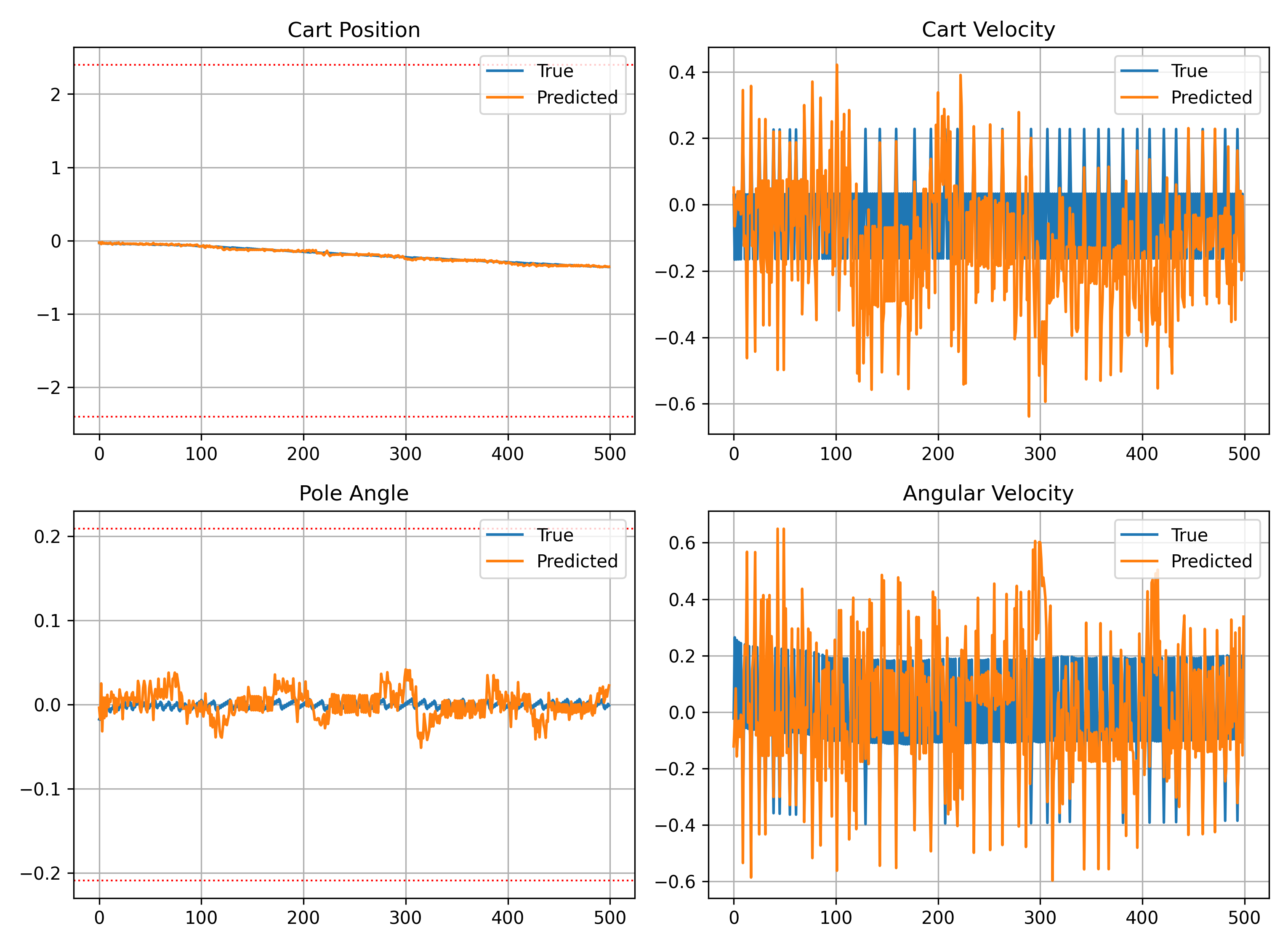}
    \caption{DQN performance with predicted state}
    \label{fig:pred_state}
\end{figure}

We then train the second DQN agent using the same hyperparameters, but with access only to the predicted state from the learned model. During evaluation, the agent is given access solely to the estimated state for decision making. We observe that the pole is controlled effectively; however, there is slight drift in cart position away from the center, as illustrated in Figure~\ref{fig:pred_state}.

\section{Evaluation Results}
As shown in Table~\ref{RMSE}, the prediction model achieves low RMSE across all state variables, with the cart position estimated most accurately (0.24\%). Pole angle and angular velocity errors remain below 4\%, indicating reliable encoding of rotational dynamics. The highest error occurs in cart velocity (3.80\%), reflecting the challenge of inferring motion from visual input. Overall, the model demonstrates effective state reconstruction from image sequences.

\begin{table}[H]
\centering
\caption{State Prediction Error (RMSE \%)}
\scriptsize
\setlength{\tabcolsep}{4pt}
\begin{tabular}{c c c c c}
\toprule
\textbf{State} & $x$ & $\dot{x}$ & $\theta$ & $\dot{\theta}$ \\
\midrule
\textbf{RMSE (\%)} & 0.24 & 3.80 & 3.89 & 3.28 \\
\bottomrule
\end{tabular}
\label{RMSE}
\end{table}

Table~\ref{MAE} compares tracking performance of DQN agents using full versus predicted states.As expected, the full-state agent achieves minimal error, particularly in cart position (0.53\%). The predicted-state agent performs comparably on pole angle (1.19\% vs. 1.60\%) and even outperforms in angular velocity. However, its cart position error increases to 5.30\%, aligning with the estimation gap in cart velocity. We observe the model maintains stable control, validating the use of predicted states for RL in perception based control setting.

\begin{table}[]
\centering
\caption{Tracking Error (MAE \%) by RL Agents}
\scriptsize
\setlength{\tabcolsep}{4pt}
\begin{tabular}{c c c c c}
\toprule
\textbf{Agent} & $x$ & $\dot{x}$ & $\theta$ & $\dot{\theta}$ \\
\midrule
Full State RL & 0.53 & 3.99 & 1.60 & 3.90 \\
Pred State RL & 5.30 & 3.04 & 1.19 & 2.70 \\
\bottomrule
\end{tabular}
\label{MAE}
\end{table}

\section{Conclusion and Future Work}
One of the primary challenges in deploying machine learning-based control policies in real-world systems is their lack of interpretability, stemming from the black-box nature of deep models. In contrast, traditional optimal control approaches based on state estimation are both theoretically grounded and widely adopted in safety-critical applications. This work demonstrates the feasibility of bridging this gap by learning reliable and interpretable reinforcement learning (RL) policies directly from high-dimensional visual observations of dynamic systems.

A promising direction for future work is to extend the current framework using a Dynamic Autoencoder (DAE) to explicitly model the next observation frame, enabling RL training directly in the learned latent space and reducing reliance on full state prediction. Additionally, incorporating a particle filter into the inference pipeline could help account for process and observation noise, improving robustness under real-world uncertainty.

\section{Use of AI in Writing}
In this document, AI tools were used to check grammar, to ensure \LaTeX\ formatting, to check sentence coherence and to understand the implementation better. However, the methodology development, literature review, experiment design and evaluation plan were manually developed based on the review of academic sources.

\bibliography{mybib}

\end{document}